%% file: main.tex
\documentclass[10pt,twocolumn,letterpaper]{article}

\usepackage{cvpr}              

\usepackage[accsupp]{axessibility}

\usepackage{graphicx}
\usepackage{amsmath}
\usepackage{amssymb}
\usepackage{booktabs}
\usepackage{bm}

\usepackage{cuted}
\usepackage{capt-of}

\usepackage{multirow}
\usepackage{verbatim}
\usepackage{comment}
\usepackage{adjustbox}
\usepackage{wrapfig}

\usepackage{url}            
\usepackage{amsfonts}       
\usepackage{nicefrac}       
\usepackage{microtype}      
\usepackage{xcolor}         

\usepackage{array}
\usepackage{xspace}

\usepackage[pagebackref,breaklinks,colorlinks]{hyperref}

\usepackage[capitalize]{cleveref}
\crefname{section}{Sec.}{Secs.}
\Crefname{section}{Section}{Sections}
\Crefname{table}{Table}{Tables}
\crefname{table}{Tab.}{Tabs.}

\input{macros.tex}

\def\cvprPaperID{3237} 
\def\confName{CVPR}
\def\confYear{2023}

\begin{document}

\title{Ego-Body Pose Estimation via Ego-Head Pose Estimation}

\author{Jiaman Li \qquad C. Karen Liu\footnotemark[2] \qquad Jiajun Wu\footnotemark[2] \\
Stanford University\\
{\tt\small \{jiamanli,karenliu,jiajunwu\}@cs.stanford.edu}
}
\maketitle



\input{0teaser}

\begin{abstract}
\begin{NoHyper}
\footnotetext{\footnotemark[2] indicates equal contribution.}
\end{NoHyper}
\vspace{-4mm}
 Estimating 3D human motion from an egocentric video sequence plays a critical role in human behavior understanding and has various applications in VR/AR. However, naively learning a mapping between egocentric videos and human motions is challenging, because the user's body is often unobserved by the front-facing camera placed on the head of the user. In addition, collecting large-scale, high-quality datasets with paired egocentric videos and 3D human motions requires accurate motion capture devices, which often limit the variety of scenes in the videos to lab-like environments. To eliminate the need for paired egocentric video and human motions, we propose a new method, Ego-Body Pose Estimation via Ego-Head Pose Estimation (\model), which decomposes the problem into two stages, connected by the head motion as an intermediate representation. \model first integrates SLAM and a learning approach to estimate accurate head motion. Subsequently, leveraging the estimated head pose as input, \model utilizes conditional diffusion to generate multiple plausible full-body motions. This disentanglement of head and body pose eliminates the need for training datasets with paired egocentric videos and 3D human motion, enabling us to leverage large-scale egocentric video datasets and motion capture datasets separately. Moreover, for systematic benchmarking, we develop a synthetic dataset, AMASS-Replica-Ego-Syn (ARES), with paired egocentric videos and human motion. On both ARES and real data, our \model model performs significantly better than the current state-of-the-art methods. 
\end{abstract}


\input{1intro}
\input{2related}

\input{3method}

\input{4experiment}

\input{5conclusion}

\input{6acknowledgement}

{\small
\bibliographystyle{ieee_fullname}
\bibliography{egbib}
}

\end{document}

%% file: macros.tex
\newcolumntype{L}[1]{>{\raggedright\let\newline\\\arraybackslash\hspace{0pt}}m{#1}}
\newcolumntype{C}[1]{>{\centering\let\newline\\\arraybackslash\hspace{0pt}}m{#1}}
\newcolumntype{R}[1]{>{\raggedleft\let\newline\\\arraybackslash\hspace{0pt}}m{#1}}

\newcommand{\eqn}[1]{Equation~\ref{#1}}
\newcommand{\fig}[1]{Figure~\ref{#1}}

\newcommand{\ignore}[1]{}

\makeatletter
\DeclareRobustCommand\onedot{\futurelet\@let@token\@onedot}
\def\@onedot{\ifx\@let@token.\else.\null\fi\xspace}

\def\eg{e.g\onedot}

\def\etal{et al\onedot}

\makeatother

\definecolor{MyDarkBlue}{rgb}{0,0.08,1}
\definecolor{MyDarkGreen}{rgb}{0.02,0.6,0.02}
\definecolor{MyDarkRed}{rgb}{0.8,0.02,0.02}
\definecolor{MyDarkOrange}{rgb}{0.40,0.2,0.02}
\definecolor{MyPurple}{RGB}{111,0,255}
\definecolor{MyRed}{rgb}{1.0,0.0,0.0}
\definecolor{MyGold}{rgb}{0.75,0.6,0.12}
\definecolor{MyDarkgray}{rgb}{0.66, 0.66, 0.66}

\definecolor{turquoise}{cmyk}{0.65,0,0.1,0.3}

\newcommand{\model}{EgoEgo\xspace}



%% file: 0teaser.tex
\begin{strip}\centering
\vspace{-40px}
\captionsetup{type=figure}
\includegraphics[width=\textwidth]{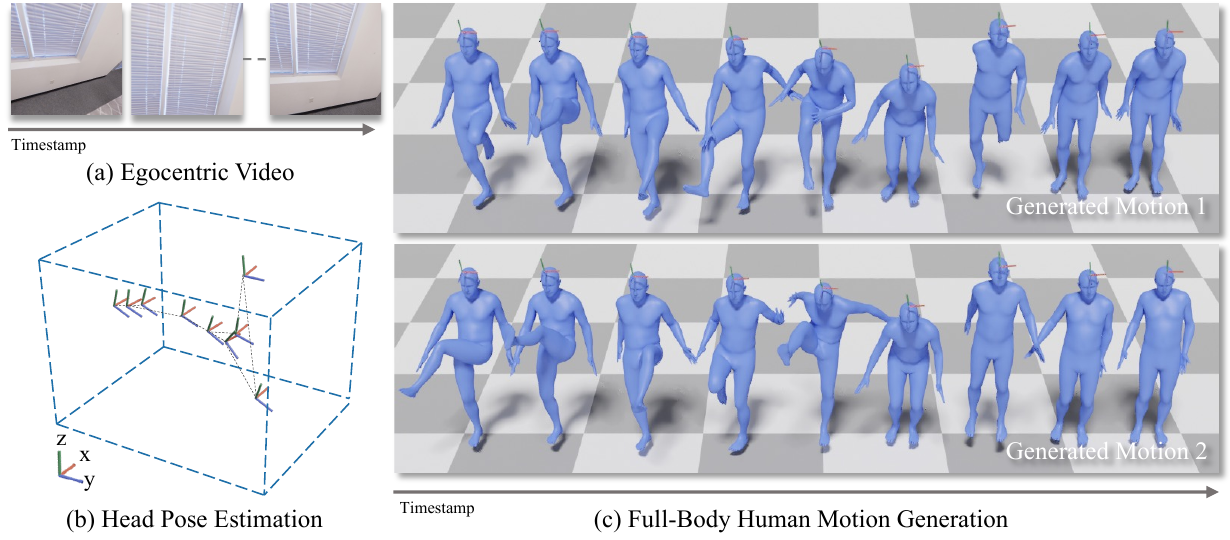}
\vspace{-20px}
\captionof{figure}{Taking egocentric video (a) as input, our approach first predicts head pose (b); it then estimates multiple plausible full-body human motions (c) from the predicted head pose. This motion sequence shows a human kicking and jumping in place. Please refer to the supplementary video on our \href{https://lijiaman.github.io/projects/egoego/}{project page} for a complete motion sequence visualization. 
\label{figure1}}
\vspace{-5px}
\end{strip}

%% file: 1intro.tex
\vspace{-5mm}
\section{Introduction}
\vspace{-1mm}
Estimating 3D human motion from an egocentric video, which records the environment viewed from the first-person perspective with a front-facing monocular camera, is critical to applications in VR/AR. However, naively learning a mapping between egocentric videos and full-body human motions is challenging for two reasons. First, modeling this complex relationship is difficult; unlike reconstruction motion from third-person videos, the human body is often out of view of an egocentric video. Second, learning this mapping requires a large-scale, diverse dataset containing paired egocentric videos and the corresponding 3D human poses. Creating such a dataset requires meticulous instrumentation for data acquisition, and unfortunately, such a dataset does not currently exist. As such, existing works have only worked on small-scale datasets with limited motion and scene diversity~\cite{yuan20183d, yuan2019ego,luo2021dynamics,HPS,zheng2022gimo}.

We introduce a generalized and robust method, \textbf{\model}, to estimate full-body human motions from only egocentric video for diverse scenarios. Our key idea is to use head motion as an intermediate representation to decompose the problem into two stages: head motion estimation from the input egocentric video and full-body motion estimation from the estimated head motion. For most day-to-day activities, humans have an extraordinary ability to stabilize the head such that it aligns with the center of mass of the body \cite{keshner1988motor}, which makes head motion an excellent feature for full-body motion estimation. More importantly, the decomposition of our method removes the need to learn from paired egocentric videos and human poses, enabling learning from a combination of large-scale, single-modality datasets (\eg, datasets with egocentric videos or 3D human poses only), which are commonly and readily available.

The first stage, estimating the head pose from an egocentric video, resembles the localization problem. However, directly applying the state-of-the-art monocular SLAM methods~\cite{teed2021droid} yields unsatisfactory results, due to the unknown gravity direction and the scaling difference between the estimated space and the real 3D world. We propose a hybrid solution that leverages SLAM and learned transformer-based models to achieve significantly more accurate head motion estimation from egocentric video. 
In the second stage, we generate the full-body motion based on a diffusion model conditioned on the predicted head pose. Finally, to evaluate our method and train other baselines, we build a large-scale synthetic dataset with paired egocentric videos and 3D human motions, which can also be useful for future work on visuomotor skill learning and sim-to-real transfer.

Our work makes four main contributions. First, we propose a decomposition paradigm, \textbf{EgoEgo}, to decouple the problem of motion estimation from egocentric video into two stages: ego-head pose estimation, and ego-body pose estimation conditioned on the head pose. The decomposition lets us learn each component separately, eliminating the need for a large-scale dataset with two paired modalities. Second, we develop a hybrid approach for ego-head pose estimation, integrating the results of monocular SLAM and learning. Third, we propose a conditional diffusion model to generate full-body poses conditioned on the head pose. Finally, we contribute a large-scale synthetic dataset with both egocentric videos and 3D human motions as a test bed to benchmark different approaches and showcase that our method outperforms the baselines by a large margin.

%% file: 2related.tex
\section{Related Work}
\paragraph{Motion Estimation from Third-person Video.}
3D pose estimation from images and videos in third-person view has been extensively studied in recent years. There are mainly two typical categories in this direction. One is to regress joint positions directly from images and videos~\cite{zhou2016sparseness,pavlakos2017coarse,mehta2017vnect,tung2017self}. The other adopts parametric human body model~\cite{SMPL:2015} to estimate body model parameters from images or videos~\cite{kanazawa2018end,kolotouros2019learning,kocabas2021pare,kocabas2021spec,kocabas2020vibe,choi2021beyond,luo20203d}. 
And recently, learned motion prior is applied to address the issues of jitters, lack of global trajectory, and missing joints or frames~\cite{yuan2021glamr, rempe2021humor, li2021task}. Moreover, physical constraints are enforced in motion estimation from videos~\cite{yuan2021simpoe,xie2021physics}. 
In contrast to third-person videos, where the full body can be seen, body joints are mostly not visible in an egocentric video, which poses a significant challenge for the problem. Although the body joints are unobserved from egocentric views, the visual information of how the environment changes provides a strong signal to infer how the head moves. In this work, we propose to use the head pose as an intermediate representation to bridge the egocentric video and full-body motions.
\vspace{-4mm}
\paragraph{Motion Estimation from Egocentric Video.}
Growing attention is received in pose estimation from egocentric videos. Special hardware like the fisheye camera is deployed to predict full body pose from captured images~\cite{tome2020selfpose,wang2021estimating, tome2019xr,xu2019mo,jiang2021egocentric}. While body joints are usually visible in images captured with a fisheye camera, the distortion of images poses a significant challenge. Jiang \etal~\cite{jiang2017seeing} deploy a standard camera to a human chest and propose an implicit motion graph matching approach to predict full body motions from the input video. You2Me~\cite{ng2020you2me} predicts full body motions by observing the interaction pose of the second-person in the camera view. Towards a goal of estimating and forecasting physically plausible motions from a head-mounted camera, EgoPose~\cite{yuan2019ego,yuan20183d} develop a Deep-RL framework to learn a control policy to estimate current poses and forecast future poses. Follow-up work Kinpoly~\cite{luo2021dynamics} integrates kinematics and dynamics to predict physically plausible motions interacting with known objects. While their method achieves impressive results in their collected dataset, it cannot handle scenes and motions out of their data distribution. This work aims to establish a more generalized and robust framework to infer full-body motions from egocentric video only. To validate the effectiveness in more generalized scenes and motions, we also introduce an approach of synthesizing egocentric video corresponding to mocap data in diverse 3D scenes for quantitative evaluation. 
\vspace{-4mm}
\paragraph{Motion Estimation from Sparse Sensors.}
Instead of estimating motion from videos, some work explored reconstructing human motions from sensors~\cite{von2017sparse,DIP:SIGGRAPHAsia:2018,guzov23ireplica}. TransPose~\cite{yi2021transpose} proposes a real-time pipeline to predict full body motions from 6 IMU sensors, including head, torso, left/right arms, and left/right lower legs. 
Follow-up work PIP~\cite{yi2022physical} further includes a PD controller on top of the kinematic estimator to introduce physics constraints during reconstruction. TIP~\cite{jiang2022transformer} follows the same sensor setting and deploys a transformer-based model to leverage IMU sequential information effectively. Fewer sensors are investigated in LoBSTr~\cite{yang2021lobstr}. Given tracker information from 4 joints (head, left/right hands, torso), they present an RNN-based model to infer lower-body motions from past upper-body joint signals. Recent advances~\cite{dittadi2021full,aliakbarian2022flag,winkler2022questsim,jiang2022avatarposer} further relax the input constraints to head and hand signals only. 
In this work, we do not rely on any observations from inertial sensors. Instead, we aim to develop a solution with egocentric video input only.

%% file: 3method.tex
\section{Method}

Our method, \model, estimates 3D human motion from a monocular egocentric video sequence. As shown in \fig{fig:overview}, our key idea is to leverage \emph{head motion}: first estimating head motion from egocentric video, and then estimating full body motion from head motion. We show that head motion is an excellent feature for full-body motion estimation and a compact, intermediate representation that reduces the challenge into two much simpler sub-problems. Such a disentanglement also allows us to leverage a large-scale egocentric video dataset with head motion (but no full body motion) in stage one, and a separate 3D human motion dataset (but no egocentric videos) in stage two. 

\vspace{-4mm}
\paragraph{Notations.} We denote full body motion as $\bm{X} \in \mathbb{R}^{T \times D}$ and egocentric images captured from a front-facing, head-mounted camera as $\bm{I} \in \mathbb{R}^{T \times h \times w \times 3}$, where $T$ is the sequence length, $D$ is the dimension of the pose state, and $h \times w$ is the size of an image. We introduce head motion $\bm{H} \in \mathbb{R}^{T \times D'}$ as an intermediate representation to bridge the input egocentric video and the output human motions, where $D'$ is the dimension of the head pose. 

\subsection{Head Pose Estimation from Egocentric Video}
Estimating the head motion from an egocentric video can be viewed as a camera localization problem. However, we observed three issues that prevent us from directly applying the state-of-the-art monocular SLAM method \cite{teed2021droid} to our problem. First, the gravity direction of the estimated head pose is unknown. Thus, the results cannot be directly fed to the full-body motion estimator, since it expects the head pose expressed in a coordinate frame where the gravity direction is $[0, 0, -1]^T$. Second, the estimated translation by monocular SLAM is not to scale when compared with the distance in the real world. Third, monocular SLAM tends to be less accurate in estimating relative head rotation than translation.

\begin{figure*}[t!]
\begin{center}
\vspace{-5mm}
\includegraphics[width=\textwidth]{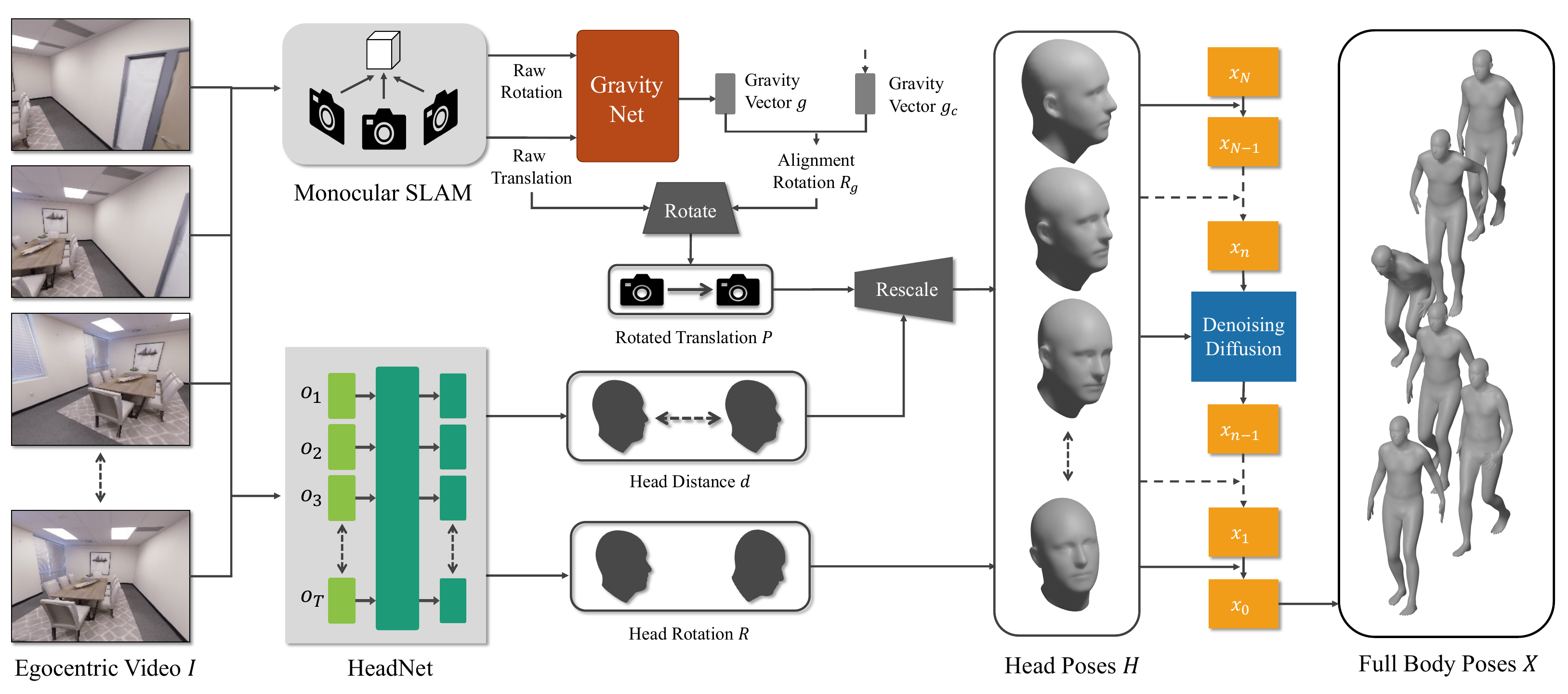}
\end{center}
\vspace{-6mm}
\caption{Overview of EgoEgo. The model first takes an egocentric video as input and predicts the head pose with a hybrid approach that combines monocular SLAM and the learned GravityNet and HeadNet. The predicted head pose is then fed to a conditional diffusion model to generate the full-body pose.}
\label{fig:overview}
\vspace{-1mm}
\end{figure*}

Based on these observations, we propose a hybrid method that leverages SLAM and learned models to achieve more accurate head pose estimation than the state-of-the-art SLAM alone. First, we develop a transformer-based model GravityNet to estimate the gravity direction from the rotation and the translation trajectories computed by SLAM. We rotate the SLAM translation by aligning the estimated gravity direction with the real gravity direction $[0, 0, -1]^{T}$ in the 3D world. Moreover, from the optical flow features extracted from the egocentric video, our method learns a model, HeadNet, to estimate head rotations and translation distance. The predicted translation distance of HeadNet is used to re-scale the translation estimated by SLAM. The predicted head rotation by HeadNet is directly used to replace the rotation estimated by SLAM. Figure \ref{fig:overview} summarizes our process to generate head poses.

\vspace{-4mm}
\paragraph{Monocular SLAM.}
We adopt DROID-SLAM~\cite{teed2021droid} to estimate camera trajectory from egocentric videos. DROID-SLAM~\cite{teed2021droid} is a learning-based method to estimate camera pose trajectory and reconstruct the 3D map of the environment simultaneously. By a design of recurrent iterative updates to camera pose and depth, it showcases superior and more robust results compared to prior SLAM systems~\cite{campos2021orb}. For more details, please refer to ~\cite{teed2021droid}. 

\begin{figure}[t!]
\begin{center}
\vspace{-2mm}
\includegraphics[width=0.45\textwidth]{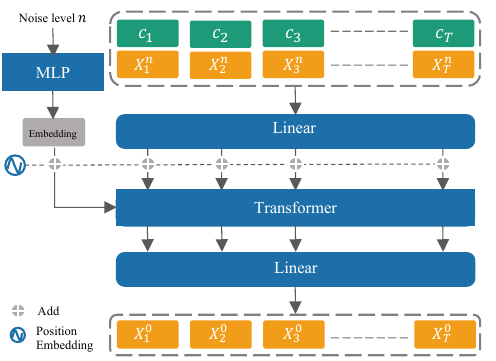}
\end{center}
\vspace{-6mm}
  \caption{Model architecture of the denoising network in a single step of the reverse diffusion process.}
  \label{fig:full_motion_estimator}
\vspace{-4mm}
\end{figure}

\vspace{-4mm}
\paragraph{Gravity Direction Estimation.}
We introduce GravityNet to predict gravity direction $\bm{g} \in \mathbb{R}^3$ from a sequence of head poses $\bm{\hat{h}}_{1}, \bm{\hat{h}}_{2}, ..., \bm{\hat{h}}_{T}$. The gravity direction $\bm{g}$ is represented by a unit vector. The head poses input $\bm{\hat{h}}_{t}$ consists of a 3D head translation, a head rotation represented by a continuous 6D rotation vector~\cite{zhou2019continuity}, head translation difference, and head rotation difference computed by $\bm{O}_{t-1}^{-1}\bm{O}_{t}$ where $\bm{O}_{t}$ denotes the head rotation matrix at time step $t$. We adopt a transformer-based architecture~\cite{vaswani2017attention} consisting of two self-attention blocks, each of which has a multi-head attention layer followed by a position-wise feed-forward layer. We take the first output of the transformer and feed it to an MLP to predict the gravity direction $\bm{g}$. We train our GravityNet on the large-scale motion capture dataset AMASS~\cite{mahmood2019amass}. However, the motion sequences in AMASS have the correct gravity direction $\bm{g}_c = [0, 0, -1]^T$. To emulate the distribution of the predicted head poses from monocular SLAM, we apply a random scale and a random rotation to the head poses in each AMASS sequence to generate our training data for gravity estimation. $L_1$ loss for the gravity vector is used during training. Based on the prediction of GravityNet, we compute the rotation matrix $\bm{R}_g$ to align the prediction $\bm{g}$ and $\bm{g}_c$. Then we apply $\bm{R}_g$ to the SLAM translation denoted as $\bm{\hat{P}}_1, \bm{\hat{P}}_2, ..., \bm{\hat{P}}_T$ and get $\bm{P}_1, \bm{P}_2, ..., \bm{P}_T$, where $\bm{P}_t = \bm{R}_g\bm{\hat{P}}_t$.     

\vspace{-4mm}
\paragraph{Head Pose Estimation.}
We propose HeadNet to predict a sequence of distance $d_1, d_2, ..., d_T$ and head rotations $\bm{R}_1, ..., \bm{R}_T$ from a sequence of optical flow features $\bm{o}_1, ..., \bm{o}_T$. The optical flow features are extracted by a pre-trained ResNet-18~\cite{he2016deep}. We deploy the same model architecture as GravityNet. Since the scale from the monocular SLAM system may not be consistent with the real 3D world where human moves, we use HeadNet to predict the vector norm of the translation difference between consecutive time steps denoted as $d_1, ..., d_T$, where $d_t$ represents a scalar value. For each camera translation sequence produced by monocular SLAM and rotated by aligning gravity direction, given the camera translation trajectory $\bm{P} \in \mathbb{R}^{T \times 3}$, we calculate the distance $d_t^{s}$ between $\bm{P}_{t}$ and $\bm{P}_{t+1}$ as $d_t^{s} = ||\bm{P}_{t+1}-\bm{P}_{t}||_2$. We take the mean value for the sequence of distances as $d^{s} = \frac{1}{T} \sum_{t=1}^{T}d_{t}^{s}$. Similarly, we compute the mean of the predicted distance sequence $d = \frac{1}{T} \sum_{t=1}^{T}d_{t}$. The scale is calculated as $s = \frac{d}{d^{s}}$. We multiply scale $s$ to the predicted translation $\bm{P}$ and use $s\bm{P}$ as our global head translation results. 

The network also predicts the head angular velocity, $\bm{\omega}_1, ..., \bm{\omega}_T$, in the head frame. We integrate predicted angular velocity to generate corresponding rotations $\bm{R}_1, ..., \bm{R}_T$. During inference, we assume that the first head orientation is given and integrate the predicted head angular velocity to estimate the subsequent head orientations.

The training loss of the HeadNet is defined as: $\mathcal{L} = \mathcal{L}_{dist} + \mathcal{L}_{vel} + \mathcal{L}_{rot}$. $\mathcal{L}_{dist}$ represents the $L1$ loss for translation distance. $\mathcal{L}_{vel}$ represents the $L1$ loss for angular velocity. $\mathcal{L}_{rot}$ denotes the rotation loss $\mathcal{L}_{rot} = ||\bm{R}_{pred}\bm{R}_{gt}^{T} - \bm{I}||_{1}$ where $\bm{R}_{pred}$ represents the integrated rotation using predicted angular velocity, $\bm{R}_{gt}$ represents the ground truth rotation matrix and $\bm{I}$ represents the identity matrix.

\subsection{Full-Body Pose Estimation from Head Pose}
Predicting full-body pose from head pose is not a one-to-one mapping problem as different full-body motions may have the same head pose. 
Thus, we formulate the task using a conditional generative model. Inspired by the recent success of the diffusion model in image generation~\cite{rombach2022high}, we deploy a diffusion model to generate full-body poses conditioned on head poses. We use the formulation proposed in the denoising diffusion probabilistic model (DDPM)~\cite{ho2020denoising}, which has also been applied in some concurrent work~\cite{zhang2022motiondiffuse,kim2022flame,tevet2022human} for motion generation and motion interpolation tasks. We will first introduce our data representation and then detail the conditional diffusion model formulation.

A body pose $\bm{X}_t \in \mathbb{R}^D$ at time $t$ consists of the global joint position ($\mathbb{R}^{J \times 3}$) and global joint rotations ($\mathbb{R}^{J \times 6}$). We adopt the widely used SMPL model~\cite{SMPL:2015} as our skeleton, and the number of joints $J$ is 22. For the convenience of notation in the diffusion model, we use $\bm{x}_n$ to denote a sequence of body poses $\bm{X}_1^n, \bm{X}_2^n, ..., \bm{X}_T^n$ at noise level $n$. 

The high-level idea of the diffusion model is to design a forward diffusion process to add Gaussian noises to the original data with a known variance schedule and learn a denoising model to gradually denoise $N$ steps given a sampled $\bm{x}_N$ from a normal distribution to generate $\bm{x}_0$. 

Specifically, diffusion models consist of a forward diffusion process and a reverse diffusion process. The forward diffusion process gradually adds Gaussian noise to the original data $\bm{x}_0$. And it is formulated using a Markov chain of $N$ steps as shown in \eqn{eq:forward_diffusion}:
\begin{equation}
\label{eq:forward_diffusion}
    q(\bm{x}_{1:N}|\bm{x}_{0}) := \prod_{n=1}^{N} q(\bm{x}_{n}|\bm{x}_{n-1}).
\end{equation}
Each step is decided by a variance schedule using $\beta_{n}$ and is defined as
\begin{equation}
\label{eq:forward_diffusion_step}
    q(\bm{x}_{n}|\bm{x}_{n-1}) := \mathcal{N}(\bm{x}_{n}; \sqrt{1-\beta_{n}}\bm{x}_{n-1}, \beta_{n}\bm{I}).
\end{equation}
To generate full-body motion conditioned on the head pose, we need to reverse the diffusion process. The reverse process can be approximated as a Markov chain with a learned mean and fixed variance:   
\begin{equation}
\label{eq:reverse_diffusion_step}
    p_{\theta}(\bm{x}_{n-1}|\bm{x}_{n}, c) := \mathcal{N}(\bm{x}_{n-1}; \bm{\mu}_{\theta}(\bm{x}_n, n, c), \sigma_{n}^{2}\bm{I}).
\end{equation}
where $\bm{\theta}$ represents the parameters of a neural network, $c$ is the head conditions. The learned mean $\bm{\mu}_{\theta}(\bm{x}_n, n, c)$ (we use $\bm{\mu}_{\theta}$ in the equation for brevity) can be represented as follows where $\alpha_{n}$ and $\bar{\alpha}_{n}$ are fixed parameters, $\hat{\bm{x}}_{\theta}(\bm{x}_{n}, n, c)$ is the prediction of $\bm{x}_0$:
\begin{equation}
\label{eq:reverse_diffusion_mu}
     \bm{\mu}_{\theta} = \frac{\sqrt{\alpha_{n}}(1-\bar{\alpha}_{n-1})\bm{x}_{n}+\sqrt{\bar{\alpha}_{n-1}}(1-\alpha_{n})\hat{\bm{x}}_{\theta}(\bm{x}_{n}, n, c)}{1-\bar{\alpha}_{n}}
\end{equation}
Learning the mean can be reparameterized as learning to predict the original data $\bm{x}_0$. The training loss is defined as a reconstruction loss of $\bm{x}_{0}$:
\begin{equation}
\label{eq:loss}
     \mathcal{L} = \mathbb{E}_{\bm{x}_0, n}||\hat{\bm{x}}_{\theta}(\bm{x}_{n}, n, c) - \bm{x}_{0}||_{1}
\end{equation}
As shown in Figure~\ref{fig:full_motion_estimator}, in denoising step $n$, we concatenate head pose condition $\bm{c}_1, ..., \bm{c}_T$ with body pose representation $\bm{X}_{1}^{n}, ..., \bm{X}_{T}^{n}$ at noise level $n$, combined with noise embedding as input to a transformer model, and estimate $\bm{x}_{0}$. 

\begin{figure*}[t!]
\begin{center}
\includegraphics[width=\textwidth]{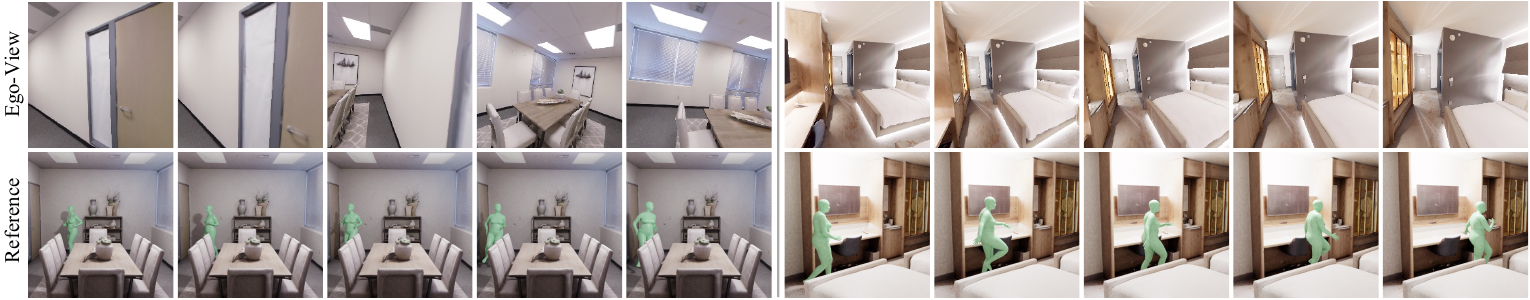}
\end{center}
\vspace{-6mm}
\caption{Illustration of our ARES Dataset. ``Ego-View'' represents the synthetic egocentric images using our proposed data generation pipeline. We also provide a third-person view reference in the second row. The left and right show two sequences from different scenes.}
\label{fig:syn_data}
\vspace{-2mm}
\end{figure*}

\subsection{Synthetic Data Generation}
Our method does not need paired training data. Still, for benchmarking purposes, we develop a way to automatically synthesize a large-scale dataset with various paired egocentric videos and human motions. 
\vspace{-4mm}
\paragraph{Generate Motions in 3D Scenes.}
To generate a dataset with both egocentric video and ground truth human motions, we use a large-scale motion capture dataset AMASS~\cite{mahmood2019amass} and a 3D scene dataset Replica~\cite{straub2019replica}. We convert the scene mesh from Replica to the signed distance field (SDF) for the penetration calculation. We divide each sequence of AMASS~\cite{mahmood2019amass} into sub-sequences with 150 frames. For each sub-sequence, based on the semantic annotation provided by Replica~\cite{straub2019replica}, we place the first pose in a random location with the feet in contact with the floor. Then we calculate penetration loss following Wang \etal~\cite{wang2021synthesizing} for each pose in this sequence. We empirically set the threshold to 2 and only keep the poses with penetration loss less than the threshold. Specifically, for human mesh $\bm{M}_t$ at time $t$ represented by a parameterized human model~\cite{SMPL:2015,SMPL-X:2019}, we denote $d_i$ as the signed distance of vertex $i$. The penetration loss is then defined as $L_{pen}^{t} = \sum_{d_i < 0} ||d_{i}||$.

\vspace{-4mm}
\paragraph{Synthesize Realistic Egocentric Images.}

The motion sequences produced by detecting penetration with 3D scenes provide the camera pose trajectories to render synthetic egocentric videos. AI Habitat~\cite{habitat19iccv,szot2021habitat} is a platform for embodied agent research that supports fast rendering given a camera trajectory and a 3D scene. We feed the head pose trajectories to the platform and synthesize realistic images in the egocentric view. We generate 1,664,616 frames with 30 fps, approximately 15 hours of motion in 18 scenes. We name the synthetic dataset AMASS-Replica-Ego-Syn (ARES) and show some examples from our synthetic dataset in Figure~\ref{fig:syn_data}. 





%% file: 4experiment.tex
\begin{table*}[t!]
\small
\begin{center}
\footnotesize{
\setlength{\tabcolsep}{4pt}
\begin{tabular}{lccccccccccccccc} 
 \toprule 
 & \multicolumn{5}{c}{ARES} & \multicolumn{5}{c}{Kinpoly-MoCap~\cite{luo2021dynamics}} & \multicolumn{5}{c}{GIMO~\cite{zheng2022gimo}} \\
 \cmidrule(lr){2-6}\cmidrule(lr){7-11}\cmidrule(lr){12-16} 
 Method     & $\mathbf{O}_{head}$ & $\mathbf{T}_{head}$ & MPJPE & Accel & FS  & $\mathbf{O}_{head}$ & $\mathbf{T}_{head}$ & MPJPE & Accel & FS  & $\mathbf{O}_{head}$ & $\mathbf{T}_{head}$ & MPJPE & Accel & FS  \\
        \midrule
        PoseReg~\cite{yuan2019ego} & 0.77 & 354.7 & 147.7 & 127.6 & 87.1 & 1.05 & 1943.9 & 160.4 & 61.8 & 10.8 & 1.51 & 1528.6 & 189.3 & 71.5 & 14.2 \\
        Kinpoly-OF~\cite{luo2021dynamics} & 0.62 & 323.4 & 141.6 & 7.3 & 4.2 & 1.33 & 2475.5 & 230.5 & 16.4 & 15.8 & 1.52 & 1739.3 & 404.2 & 21.9 & 14.4 \\
        EgoEgo (ours) & \textbf{0.20} & \textbf{148.0} & \textbf{121.1} & \textbf{6.2} & \textbf{2.7} & \textbf{0.58} & \textbf{505.1} & \textbf{125.9} & \textbf{8.0} & \textbf{1.6} & \textbf{0.67} & \textbf{356.8} & \textbf{152.1} & \textbf{10.4} & \textbf{1.9} \\
        \bottomrule
\end{tabular}
}
\end{center}
\vspace{-5mm}
\caption{Full-body motion estimation from egocentric video on ARES, Kinpoly-MoCap~\cite{luo2021dynamics}, and GIMO~\cite{zheng2022gimo}.} 
    \label{tab:complete_estimation}
\vspace{-4mm}
\end{table*}

\section{Experiments}
We evaluate and compare our method to baselines on five commonly used metrics for human motion reconstruction, in addition to the human perception studies. We also conduct ablation studies to analyze the performance of each stage of our method, as well as the design choices in our model.


\subsection{Datasets and Evaluation Metrics}

\noindent\textbf{AMASS-Replica-Ego-Syn (ARES)} is our synthetic dataset which contains synthetic egocentric videos and ground truth motions. ARES contains about 15 hours of motion across 18 scenes. We remove 5 scenes from training as unseen scenes. The training dataset consists of about 1.2M frames in 13 different scenes. The testing dataset contains 34, 850 frames from 5 unseen scenes.  


\vspace{1mm}\noindent\textbf{AMASS}~\cite{mahmood2019amass} is a large-scale motion capture dataset with about 45 hours of diverse motions. We split training and testing data following HuMoR~\cite{rempe2021humor}. 


\vspace{1mm}\noindent\textbf{Kinpoly-MoCap}~\cite{luo2021dynamics} consists of egocentric videos captured using a head-mounted camera and corresponding 3D motions captured with motion capture devices. The total motion is about 80 minutes long. Since it uses motion capture devices, the egocentric video is constrained to a single lab scene. 


\vspace{1mm}\noindent\textbf{Kinpoly-RealWorld}~\cite{luo2021dynamics} contains paired egocentric videos and head poses captured using iPhone ARKit. Unlike Kinpoly-MoCap which is captured in a lab scene, Kinpoly-RealWorld provides in-the-wild egocentric videos. 


\vspace{1mm}\noindent\textbf{GIMO}~\cite{zheng2022gimo} consists of egocentric video, eye gaze, 3D motions, and scanned 3D scenes. This dataset is collected using Hololens, iPhone 12, and IMU-based motion capture suits to study motion prediction tasks guided by eye gaze. We use 15 scenes for training and 4 scenes for testing. 

\vspace{-4mm}
\paragraph{Evaluation Metrics.}
\begin{itemize}
   \vspace{-2mm}
    \item \textbf{Head Orientation Error $\left( \mathbf{O}_{head} \right)$} computes the Frobenius norm of the difference between the $3 \times 3$ rotation matrix $||\bm{R}_{pred}\bm{R}_{gt}^{-1} - \bm{I}||_{2}$, where $\bm{R}_{pred}$ is the predicted head rotation matrix and $\bm{R}_{gt}$ is the ground truth head rotation matrix. 
    \vspace{-2mm}
    \item \textbf{Head Translation Error $\left( \mathbf{T}_{head} \right)$} is computed by taking the mean Euclidean distance of two trajectories. We use this metric to measure the head joint translation errors in millimeters $\left(mm\right)$.
    \vspace{-2mm}
    \item \textbf{MPJPE} represents mean per-joint position errors in millimeters $\left(mm\right)$.  
    \vspace{-2mm}
    \item \textbf{Accel} represents the difference of acceleration between predicted joint positions and ground truth joint positions measured in $\left(mm/s^2\right)$. 
    \vspace{-2mm}
    \item \textbf{FS} represents foot skating metric and is computed following NeMF~\cite{he2022nemf}. Specifically, we first project the toe and ankle joints' velocity to the $xy$ plane and compute the $L{1}$ norm of the projected velocities in each step denoted as $v_{t}$. We only accumulate the horizontal translation for those steps that have a height $h_{t}$ lower than a specified threshold $H$. And the metric is calculated as a mean of weighted values $v_{t}(2-2^{\frac{h_{t}}{H}})$ across the sequence and is measured in $\left(mm\right)$.      
\end{itemize}

\begin{figure*}[t!]
\begin{center}
\includegraphics[width=\textwidth]{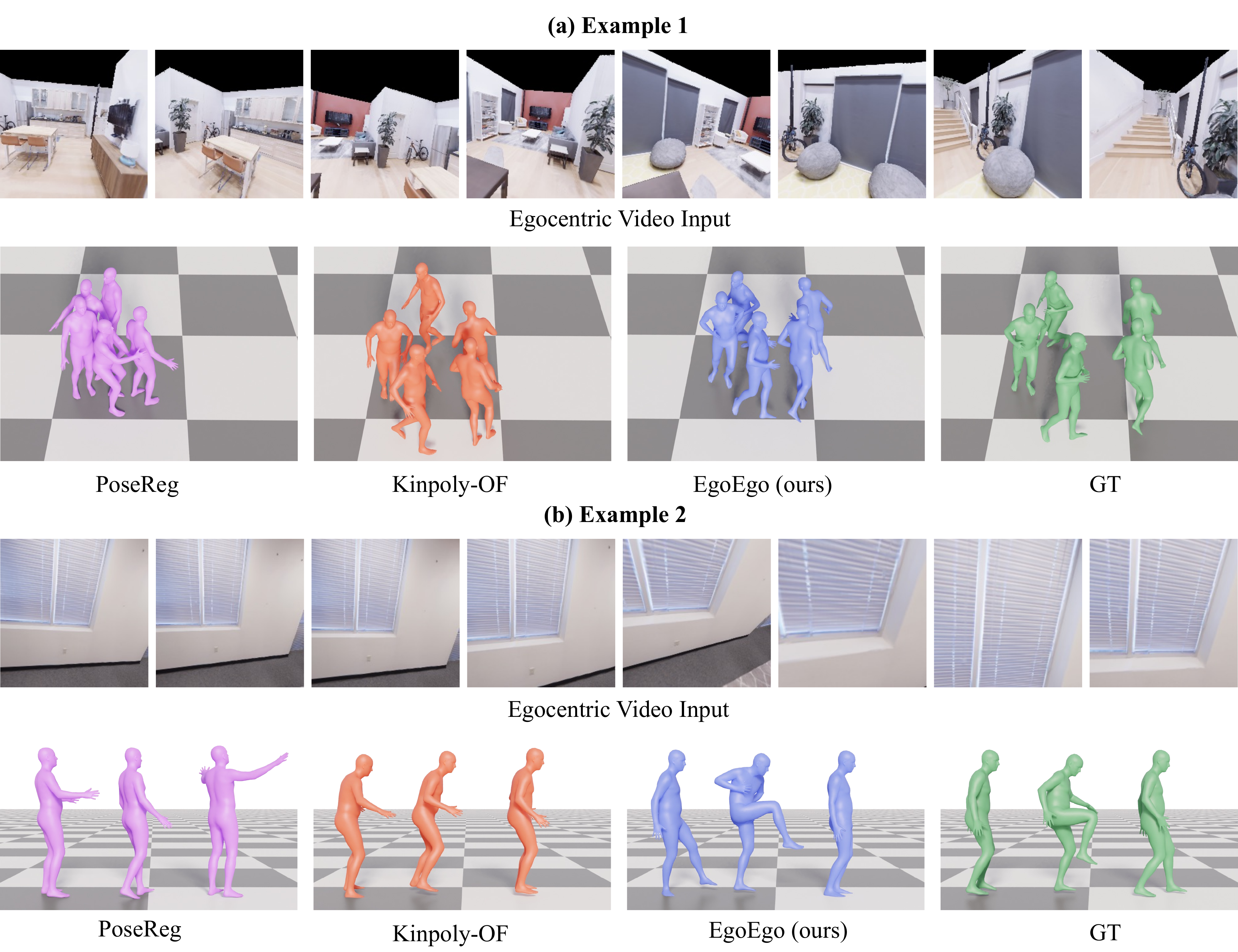}
\end{center}
\vspace{-6mm}
\caption{Qualitative results comparisons. We show the results of two egocentric videos from different scenes.}
\label{fig:qualitative_res}
\vspace{-2mm}
\end{figure*}

\subsection{Body Pose Estimation from Egocentric Video} 
\vspace{-1mm}
\paragraph{Training Data.}
We train our HeadNet using paired egocentric videos and head poses provided by ARES, Kinply-RealWorld, and GIMO. Note that the body motion in these datasets is not used for training HeadNet. Our GravityNet and the conditional diffusion model are both trained on AMASS. For the baselines below, we train them using paired egocentric videos and ground truth motions in ARES. 
\vspace{-4mm}
\paragraph{Baselines.} 
We compare our approach with two baselines PoseReg~\cite{yuan2019ego} and Kinpoly~\cite{luo2021dynamics}. \textbf{PoseReg}~\cite{yuan2019ego} takes a sequence of optical flow features as input and uses an LSTM model to predict the pose state at each time step. The pose state consists of root translation, root orientation, joint rotations, and corresponding velocities including root linear velocity and angular velocities of all the joints. \textbf{Kinpoly-OF}~\cite{luo2021dynamics} proposes a per-step regression model to estimate full body motion from optical flow features. Because our problem only allows for egocentric video as input, we choose the option of Kinpoly which only has optical flow features as input, without relying on ground truth head poses and action labels that depend on additional knowledge.

\vspace{-5mm}
\paragraph{Results.}
We compare the complete pipeline of EgoEgo with baseline methods PoseReg~\cite{yuan2019ego} and Kinpoly-OF~\cite{luo2021dynamics} on ARES, Kinpoly-MoCap~\cite{luo2021dynamics} and GIMO~\cite{zheng2022gimo}, as shown in Table~\ref{tab:complete_estimation}. We show that our EgoEgo outperforms all the baselines by a large margin on all three datasets. We show qualitative results in Figure~\ref{fig:qualitative_res}. Our generated motions better preserve the root trajectories. And our approach can also generate more dynamic and realistic motions compared to the baselines.

\subsection{Head Pose Estimation from Egocentric Video}
\vspace{-1mm}
\paragraph{Baselines.}
 We compare our hybrid approach with the prediction results of DROID-SLAM~\cite{teed2021droid}. For a fair comparison, we apply a rotation to the SLAM trajectory by aligning the first predicted head pose of SLAM with the ground truth head pose. We train our GravityNet on AMASS training split. As for HeadNet, we train on ARES, Kinpoly-RealWorld, and GIMO separately for the evaluation of different datasets.

\vspace{-4mm}
\paragraph{Results.}
We evaluate the head pose estimation on the three datasets as shown in Table~\ref{tab:head_pose_estimation}. We show more accurate head rotation prediction results on ARES and comparable results on real-captured data. As the real-captured data is limited in scale (Kinpoly-RealWorld contains 20 minutes of training videos and GIMO contains 30 minutes of training videos), we believe the head rotation prediction can be further improved by future developments of large-scale real-captured datasets with head poses. Overall, our hybrid approach combines the accurate rotation prediction from HeadNet and re-scaled translation of gravity-aligned SLAM results, and produces more accurate head pose estimation results as input to the second stage.  

\begin{table}[!t]
    \centering
    \small
      \begin{tabular}{lcccccccc}
        \toprule
        
         & \multicolumn{2}{c}{DROID-SLAM~\cite{teed2021droid}} & 
        \multicolumn{2}{c}{Ours} \\
        
        \cmidrule(lr){2-3}\cmidrule(lr){4-5} 
        
        & $\mathbf{O}_{head}$  & $\mathbf{T}_{head}$ & $\mathbf{O}_{head}$  & $\mathbf{T}_{head}$  \\
        
        \midrule
        
        ARES & 0.62 & 411.3 & \textbf{0.23} & \textbf{176.5}  \\
        Kinpoly-MoCap & \textbf{0.55} & 1290.8 & 0.58 & \textbf{487.8} \\
        GIMO & \textbf{0.67} & 865.4 & 0.68 & \textbf{304.7} \\
        \bottomrule
       
      \end{tabular}
    \vspace{-2mm}
     \caption{Head pose estimation on test sets. 
     }
    \label{tab:head_pose_estimation}
    \vspace{-4mm}
\end{table}

\subsection{Body Pose Estimation from the Head Pose}
\vspace{-1mm}
\paragraph{Baselines.}
We compare our conditional diffusion model for full-body pose estimation with two baselines AvatarPoser~\cite{jiang2022avatarposer} and Kinpoly-Head~\cite{luo2021dynamics}. \textbf{AvatarPoser}~\cite{jiang2022avatarposer} takes both head and hand pose as input to predict full body motion. We remove the hand poses from the input and modified it to a setting with head pose input only. \textbf{Kinpoly-Head}~\cite{luo2021dynamics} is our modified variant of the Kinpoly model that only takes the head pose as input. Both the baselines and our method are trained on the training split of AMASS with high-quality motion capture data.

\begin{table}[t!]
\small
\begin{center}
\footnotesize{
\setlength{\tabcolsep}{5pt}
\begin{tabular}{lccccc}
 \toprule 
 Method     & $\mathbf{O}_{head}$ & $\mathbf{T}_{head}$ & MPJPE & Accel & FS  \\
        \midrule
        AvatarPoser~\cite{jiang2022avatarposer} & 0.19 & \textbf{28.6} & 124.7 & 16.1 & 18.8 \\
        Kinpoly-Head~\cite{luo2021dynamics} & 0.19 & 87.8 & 110.9 & 11.4 & 11.2  \\
        Diffusion Model (ours) & \textbf{0.04} & 36.7 & \textbf{109.0} & \textbf{10.5} & \textbf{4.4} \\
        \bottomrule
\end{tabular}
}
\end{center}
\vspace{-6mm}
\caption{Full-body motion estimation from GT head pose on AMASS testing dataset~\cite{luo2021dynamics}. } 
    \label{tab:stage2_estimation_amass}
\vspace{-2mm}
\end{table}

\vspace{-4mm}
\paragraph{Results.}
We evaluate the baselines and our method on the AMASS test set, as shown in Table~\ref{tab:stage2_estimation_amass}. Since our model is generative, there are multiple plausible predictions from the same head pose input. For a quantitative comparison, we generate 200 samples for each head pose input and use the one with the smallest MPJPE as our result.

\subsection{Ablation Studies} 
\vspace{-1mm}
\paragraph{Effects of Components in Head Pose Estimation.}
We study the effects of each component in head pose estimation in Table~\ref{tab:ablation_head_pose_hybrid}. We showcase that the rotation for aligning gravity direction and the learned scale are both effective to improve head translation results. 

\begin{table}[!t]
    \centering
    \small
    \setlength{\tabcolsep}{2.7pt}
      \begin{tabular}{lcccccccc}
        \toprule
        
         & \multicolumn{2}{c}{ARES} & 
        \multicolumn{2}{c}{Kinpoly-MoCap} & 
        \multicolumn{2}{c}{GIMO} \\
        
        \cmidrule(lr){2-3}\cmidrule(lr){4-5}\cmidrule(lr){6-7}  
        
        & $\mathbf{O}_{head}$  & $\mathbf{T}_{head}$ & $\mathbf{O}_{head}$  & $\mathbf{T}_{head}$ & $\mathbf{O}_{head}$  & $\mathbf{T}_{head}$  \\
        
        \midrule
        SLAM & 0.62 & 411.33 & \textbf{0.55} & 1290.82 & \textbf{0.67} & 865.41 \\
        SLAM+S & 0.62 & 325.95 & 0.55 & 643.45 & 0.67 & 569.48 \\
        SLAM+S+G & 0.62 & 176.54 & 0.55 & 487.77 & 0.67 & 304.74 \\
        Full model & \textbf{0.23} & \textbf{176.54} & 0.58 & \textbf{487.77} & 0.68 & \textbf{304.74}  \\
        \bottomrule
       
      \end{tabular}
    \vspace{-2mm}
     \caption{Ablation study for the components in head pose estimation. S represents the scale predicted by HeadNet, and G represents GravityNet.}
    \label{tab:ablation_head_pose_hybrid}
    \vspace{-4mm}
\end{table}

\vspace{-4mm}
\paragraph{Effects of Head Pose in Full-Body Pose Estimation.}
We compare the full-body pose estimation results that take our predicted head poses and the ground truth head poses as input. Table~\ref{tab:ablation_head_pose_gt} shows that the ground truth head poses significantly improve the full body pose estimation, indicating that by developing methods that predict more accurate head pose, the full body pose estimation can be further improved.     


\subsection{Human Perceptual Study}
We also conduct two human perceptual studies as part of the evaluation. The first is to evaluate the quality of predicted full-body motion from egocentric video, the second is to evaluate the quality of predicted full-body motion from ground truth head poses. In both studies, we compare four types of motion: results from our \model and from two baselines, as well as the ground truth. For the first human study, each time, users are presented with two motions and an egocentric video, and asked to select which one is more plausible. For the second human study, users are presented with two motions and asked to select which one looks more natural and realistic. Because there are 10 examples and the two motions can be from four sources, we have 60 questions for each study. Each question was answered by 20 Amazon Mechanical Turk workers.

\begin{figure}[t!]
\begin{center}
\includegraphics[width=\linewidth]{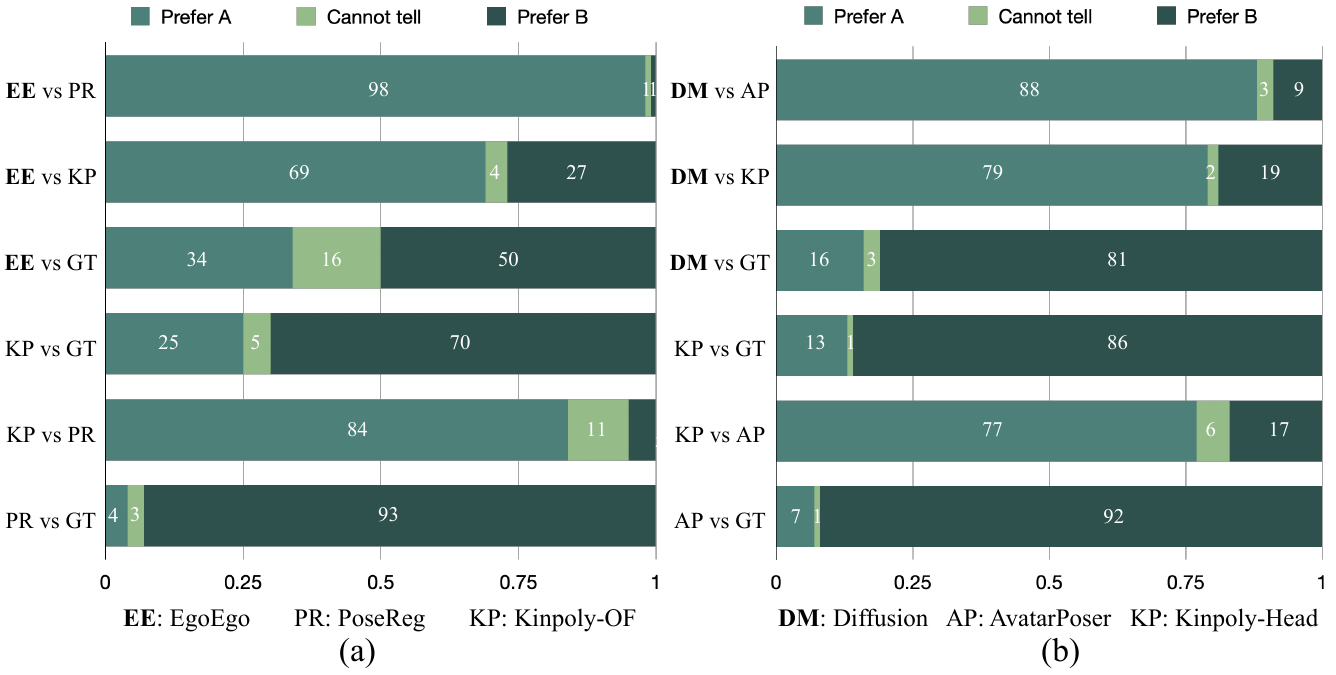}
\end{center}
\vspace{-6mm}
  \caption{Results of human perceptual studies. The numbers shown in the chart represent \%.}
  \label{fig:human_study}
\vspace{-5mm}
\end{figure}

As shown in Figure~\ref{fig:human_study}(a), for full-body estimation from egocentric video, our results are preferred by 98\% and 69\% of workers when compared to the baselines. Also, when compared with the ground truth, 34\% of the responses prefer our results (note that a perfect output would achieve 50\%), suggesting that people cannot easily distinguish our results from ground truth motions. As shown in Figure~\ref{fig:human_study}(b), for full-body estimation from head poses, our results are preferred by 88\% and 79\% of workers when compared to the baselines.

\begin{table}[t!]
    \centering
    \small
    \setlength{\tabcolsep}{2.7pt}
      \begin{tabular}{lcccccc}
        \toprule
        
         & \multicolumn{2}{c}{ARES} & 
        \multicolumn{2}{c}{Kinpoly-MoCap} & 
        \multicolumn{2}{c}{GIMO} \\
        
        \cmidrule(lr){2-3}\cmidrule(lr){4-5}\cmidrule(lr){6-7}  
        
        & EE  & EE w/ GT & EE  & EE w/ GT &  EE  & EE w/ GT  \\
        
        \midrule
        $\mathbf{O}_{head}$ & 0.20 & \textbf{0.04} & 0.58 & \textbf{0.03} & 0.67 & \textbf{0.06} \\
        $\mathbf{T}_{head}$ & 148.0 & \textbf{29.1} & 505.1 & \textbf{60.7} & 356.8 & \textbf{66.0}  \\
        MPJPE & 121.1 & \textbf{105.6} & 125.9 & \textbf{76.0} & 152.1 & \textbf{125.7} \\
        Accel & \textbf{6.2} & 6.3 & 8.0 & \textbf{7.2} & 10.4 & \textbf{10.2}  \\
        FS & \textbf{2.7} & 3.0 & \textbf{1.6} & 2.5 & 1.9 & \textbf{1.7}  \\
        \bottomrule
       
      \end{tabular}
    \vspace{-2mm}
     \caption{Ablation study for the effects of head pose to full-body pose estimation. EE represents EgoEgo. GT represents ground truth head poses.}
    \label{tab:ablation_head_pose_gt}
    \vspace{-2mm}
\end{table}

%% file: 5conclusion.tex
\vspace{-3mm}
\section{Conclusion}
We presented a generalized framework to estimate full-body motions from egocentric video. The key is to decompose the problem into two stages. We predicted the head pose from an egocentric video and fed the output from the first stage to estimate full-body motions in the second stage. In addition, we developed a hybrid solution to produce more accurate head poses on top of monocular SLAM. We also proposed a conditional diffusion model to generate diverse high-quality full-body motions from predicted head poses. To benchmark different methods in a large-scale dataset, we proposed a data generation pipeline to synthesize a large-scale dataset with paired egocentric videos and 3D human motions. We showcased superior results on both the synthetic and the real-captured dataset compared to prior work. 


%% file: 6acknowledgement.tex
\vspace{-5mm}
\paragraph{Acknowledgement.}
We thank Zhengfei Kuang for the help with visualizations and Jonathan Tseng for discussions about the diffusion model. This work is in part supported by ONR MURI N00014-22-1-2740, NSF CCRI 2120095, the Toyota Research Institute (TRI), the Stanford Institute for Human-Centered AI (HAI), Innopeak, Meta, and Samsung.